\documentclass[letterpaper, 10 pt, conference]{ieeeconf}  

\usepackage[utf8]{inputenc}
\usepackage{blindtext}
\usepackage{comment}
\usepackage{todonotes} 
\usepackage{url}
\usepackage{subcaption}
\usepackage{siunitx}

\IEEEoverridecommandlockouts                              

\title{\LARGE \bf
Extraction and Assessment of Naturalistic Human Driving Trajectories from Infrastructure Camera and Radar Sensors%
}

\author{Dominik Notz$^{1}$ 
\and 
Felix Becker$^{2}$
\and
Thomas Kühbeck$^{1}$
\and
Daniel Watzenig$^{3}$
\thanks{$^{1}$Dominik Notz ({\tt\footnotesize dominik.notz@bmw.de}) and
Thomas Kühbeck ({\tt\footnotesize thomas.kuehbeck@bmw.de})
are with BMW of North America, 2606 Bayshore Pkwy, Mountain View, CA 94043, USA.}%
\thanks{$^{2}$Felix Becker ({\tt\footnotesize felix.becker@mailbox.org}) was an intern at BMW of North America, 2606 Bayshore Pkwy, Mountain View, CA 94043, USA.}%
\thanks{$^{3}$Daniel Watzenig ({\tt\footnotesize daniel.watzenig@tugraz.at}) is with the Institute of Automation and Control at Graz University of Technology, Inffeldgasse 21/B/I, 8010 Graz, Austria.}%
}

\begin{document}

\maketitle
\thispagestyle{empty}
\pagestyle{empty}

%
%
%
%

\begin{abstract}
\label{sec:0_abstract}
%
%
Collecting realistic driving trajectories is crucial for training machine learning models that imitate human driving behavior. %
Most of today's autonomous driving datasets contain only a few trajectories per location and are recorded with test vehicles that are cautiously driven by trained drivers. %
In particular in interactive scenarios such as highway merges, the test driver's behavior significantly influences other vehicles. %
This influence prevents recording the whole traffic space of human driving behavior. %
In this work, we present a novel methodology to extract trajectories of traffic objects using infrastructure sensors. %
Infrastructure sensors allow us to record a lot of data for one location and take the test drivers out of the loop. %
We develop both a hardware setup consisting of a camera and a traffic surveillance radar and a trajectory extraction algorithm. %
Our vision pipeline accurately detects objects, fuses camera and radar detections and tracks them over time. %
We improve a state-of-the-art object tracker by combining the tracking in image coordinates with a Kalman filter in road coordinates. %
We show that our sensor fusion approach successfully combines the advantages of camera and radar detections and outperforms either single sensor. %
Finally, we also evaluate the accuracy of our trajectory extraction pipeline. 
For that, we equip our test vehicle with a differential GPS sensor and use it to collect ground truth trajectories. %
With this data we compute the measurement errors. 
While we use the mean error to de-bias the trajectories, the error standard deviation is in the magnitude of the ground truth data inaccuracy. 
Hence, the extracted trajectories are not only naturalistic but also highly accurate and prove the potential of using infrastructure sensors to extract real-world trajectories. %
\end{abstract}
\section{Introduction}
\label{sec:1_intro}
%
In the last years, the field of autonomous driving (AD) has advanced a lot and the number of disengagements per mile driven have decreased significantly \cite{dmv_disengagement}. %
One reason for this progress is the abundance of available labeled data, enabling the training of sophisticated machine learning (ML) models. %
Many companies working on self-driving have recently released AD datasets recorded with their test vehicles \cite{cordts2016cityscapes, waymo_open_dataset, lyft2019, nuscenes2019, chang2019argoverse, huang2018apolloscape}. %
These datasets focus mainly on computer vision tasks (e.g. object detection and semantic segmentation) and often consist of a collection of short driving sequences. For example, the Waymo Open Dataset contains \num{1000} segments of \SI{20}{\second} driving data from various situations \cite{waymo_open_dataset}. %

%
%

Planning, which follows perception in an AD software stack, relies on accurate behavior models for other traffic objects (TOs). %
Particularly, highly interactive scenarios such as merging onto the highway or making lane changes in dense traffic can only be handled with precise behavior models. %
Such prediction models can be trained from datasets containing TO trajectories. %
Furthermore, TO trajectory datasets enable the training of smart agents that imitate human behavior. %
Smart agent models can be used in simulation and re-processing to increase their accuracy and validity \cite{notz2019methods}. %

The released AD datasets contain only a small number of TO trajectories for each location, which are insufficient to train accurate behavior models. %
In addition, the trajectories in these datasets are recorded from test vehicles. %
Test vehicles influence the human behavior either directly, e.g. through defensive driving, or indirectly, e.g. through drawing extra attention onto them. %
Hence, to increase the coverage of the traffic space of human driving behavior in a dataset we need to passively record traffic. %
In particular, infrastructure sensors seem suitable for this task \cite{notz2019methods}.  %

%
\begin{figure}
    \centering
    \includegraphics[width=1.0\columnwidth]{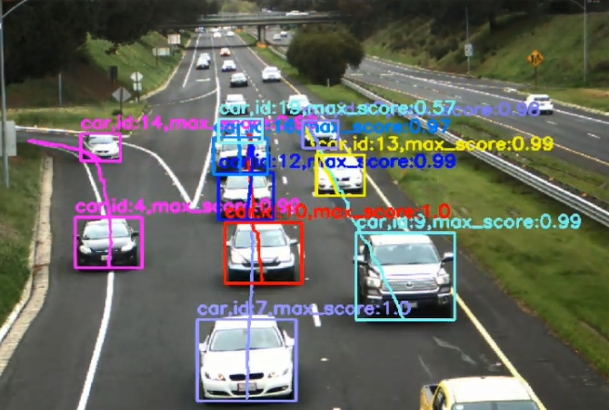}
    \caption{Visualization of detected vehicles and their past trajectories in a crowded highway merging scene. We use YOLOv3 \cite{redmon2018yolov3} to detect vehicles, which we track over time. By connecting the lower center points of all bounding boxes within on track we obtain vehicle trajectories. For each track, we extract the object class, id and maximum detection score and write them above its latest bounding box.}
    \label{fig:det_traj_image}
\end{figure}
The Next Generation Simulation (NGSIM) dataset \cite{ngsim_hwy101, ngsim_i80} contains camera traffic recordings from tall buildings from the years $2005$ and $2006$. %
However, the raw trajectories are erroneous and contain for example false positive collisions \cite{montanino2015trajectory}.
The highD dataset \cite{highDdataset} has been recorded with a drone equipped with a camera hovering over a German Autobahn. %
These recordings are better in terms of accuracy while not influencing the traffic, either. %
Nonetheless, short recording cycles due to battery limitations and effortful post-processing aggravate a recording of a multitude of large datasets. %
Besides, highD lacks information about the accuracy of the extracted trajectories. %

We develop a novel method to extract trajectories of TOs out of infrastructure sensors. %
Notably, alongside a camera, our hardware setup also consists of a traffic surveillance radar. %
Radars can directly measure the velocity of TOs and work well in adverse weather conditions \cite{medina2013evaluation}. %
We detect objects in the camera image with YOLOv3 \cite{redmon2018yolov3}, transform them via a homography into road coordinates and fuse them with radar detections. %
For object tracking, we improve a state-of-the-art intersection-over-union (IoU) tracker \cite{bochinski2017high} to better handle false negative detections. %
This improvement effectively solves the problem of identity switches and track fragmentations. %
The trajectory extraction vision pipeline runs in real-time on an Nvidia Jetson AGX. %
Figure \ref{fig:det_traj_image} shows a recorded highway merge scene with detected vehicles and their past trajectories painted into the camera image. %
Furthermore, our HD map in the OpenDrive format \cite{opendrive} allows us to add road layout and lane marking information to the trajectories. %
In comparison to existing datasets, we also assess the accuracy of the extracted trajectories. %
We equip our own vehicle with a  differential GPS (dGPS) sensor and drive repeatedly through the area covered by the sensor setup to collect ground truth data. %
We then compute the error of the extracted trajectories, including the mean and standard deviation of the error. %
We use the mean errors to de-bias the trajectories. %
The standard deviations are very small and in the same magnitude as the ground truth data inaccuracy. %
Hence, with our setup and algorithm we are able to record accurate naturalistic TO trajectories. %
Besides, we also show that our sensor fusion outperforms both camera-only and radar-only results. %

The rest of the work is structured as follows. %
In Section \ref{sec:2_rel_work}, we discuss related research and state-of-the-art vehicle trajectory datasets. %
In Section \ref{sec:3_method}, we describe our methodology for extracting vehicle trajectories out of infrastructure sensor data. %
Afterwards, in Section \ref{sec:4_experiments} we present our experiments for assessing the accuracy of the extracted trajectories. %
We present the results in Section \ref{sec:5_results}. %
Finally, we summarize our findings and give an outlook on future work. 
\section{Related Work}
\label{sec:2_rel_work}
%
In this section, we give an overview of related research. %
We take a look at datasets recorded from vehicles, drones and infrastructure sensors and we provide an overview of previous work for extracting trajectories. %

\subsection{In-Vehicle Datasets}
In $2009$ and $2010$ the European Field Operational Test (euroFOT) project \cite{benmimoun2013eurofot} gathered naturalistic driving data from over \num{1000} vehicles. %
The vehicles were equipped with various intelligent systems, e.g. GPS sensors and cameras, to record driving behavior and vehicle environment. %
The collected vehicle trajectories are not accurate enough to learn a microscopic behavior of human driving. %
More recently, several companies have collected AD datasets with their test vehicles and released them to the public \cite{cordts2016cityscapes, waymo_open_dataset, lyft2019, nuscenes2019, chang2019argoverse, huang2018apolloscape}. %
While the focus of most of these datasets is on computer vision tasks, many also contain trajectory data. %
The nuScenes dataset from Aptiv \cite{nuscenes2019} contains \num{1000} different scenes of \SI{20}{\second} duration each. %
Every \SI{0.5}{\second} a so-called keyframe has been manually annotated, yielding TO trajectories. %
The Argoverse dataset from Argo AI \cite{chang2019argoverse} contains two sub-datasets. %
Argoverse-Tracking-Beta consists of \num{100} segments of $15$ to $60$ seconds length each that have been annotated by human labelers. %
In contrast to that, Argoverse-Forecasting contains over \num{300000} \SI{5}{\second} mined, not labeled, trajectories from \SI{320}{\hour} of driving. %
%
%
In comparison to these datasets our trajectory extraction methodology is based on infrastructure sensor recordings without test drivers influencing the behavior of other TOs. %
Furthermore, infrastructure sensors allow for cost-effective and effortless recording of many trajectories from the same location. %

\subsection{Outside-of-Vehicle Datasets}
Recording TO trajectories outside of a vehicle either relies on traffic surveillance drones or fixed infrastructure sensors. %
Khan et al. \cite{khan2017unmanned} used a drone to monitor traffic and extract trajectories for high-level behavior analysis. %
The Stanford drone dataset contains human trajectories in crowded scenes, which have been collected by hovering over more than $100$ campus locations \cite{robicquet2016learning}. %
Krajewski et al. \cite{highDdataset} flew with a drone over six locations of a German Autobahn to create the highD dataset. %
This dataset consists of more than \num{100000} vehicle trajectories. %
While drones do not have a direct influence on the behavior of TOs their recordings require high efforts. %
Batteries limit single recording times to usually $10$ to $20$ minutes. %
Furthermore, recordings are only possible during sunny and windless days. %
%
%

Infrastructure sensors for traffic surveillance enable the perpetual extraction of TO trajectories. %
The NGSIM dataset \cite{ngsim_hwy101, ngsim_i80} has been recorded from tall buildings at four different locations. %
It contains naturalistic vehicle trajectories extracted from footage of several synchronized cameras from $2005$ and $2006$. %
However, its unprocessed trajectories contain errors, e.g. false positive collisions \cite{montanino2015trajectory}. %
The survey of Datondji et al. \cite{datondji2016survey} gives an overview of further camera-based vehicle monitoring approaches at intersections. %
In contrast to these datasets, we use both a camera and a radar. %
Radars directly measure the radial velocities of TOs, are independent of sunlight and work well in adverse weather conditions \cite{banerjee2018online}. %
Furthermore, we also conduct experiments to assess the accuracy of the extracted trajectories.

\subsection{Computer Vision for Trajectory Extraction}
The vision pipeline to extract trajectories from image data typically consists of object detection, coordinate transformation, tracking and post-processing \cite{ren2018learning, behbahani2019learning}. %
State-of-the-art object detectors, such as Faster R-CNN \cite{ren2015faster}, SSD \cite{liu2016ssd}, Mask R-CNN \cite{he2017mask} and YOLO \cite{redmon2016you, redmon2018yolov3}, are based on deep neural networks. %
The actual network choice depends on trade-offs between runtime and accuracy. %
If objects do not overlap in the camera image a simple foreground extraction can also be sufficient \cite{datondji2016survey}. %
The transformation of 2D bounding boxes in image coordinates into locations on the 2D road surface can be computed via a homography estimated from point correspondences \cite{chum2005geometric}. %
For the tracking of the detected objects the IoU tracker \cite{bochinski2017high} is a popular choice. %
Its extension \cite{bochinski2018extending} and Deep SORT \cite{wojke2017simple} make also use of an appearance model and several tracking algorithms can be combined into a single approach \cite{behbahani2019learning}. %
During post-processing, trajectories are smoothed. %
For non real-time smoothing, the Rauch-Tung-Striebel (RTS) smoother \cite{rauch1965maximum} is widely used. %

%
%
%
\section{Trajectory Extraction}
\label{sec:3_method}
%
%
%
Next, we describe our fully automated pipeline for the extraction of trajectories from infrastructure sensor data. %

\subsection{System Overview}
\label{subsec:systemoverview}
Our sensor setup consists of both a camera and a traffic surveillance radar. %
Cameras excel at object classification and provide good lateral object position accuracy. %
Radars yield the exact distance to objects and their radial velocity. %
Furthermore, they work well in adverse weather conditions and darkness. %
%
%
Our vision pipeline to extract TO trajectories runs in real-time on an Nvidia Jetson AGX Xavier \cite{jetsonagx}. %
We do not save the camera feed to disk but only detected objects or extracted trajectories, respectively. %
With this implicit data anonymization we comply with privacy laws and keep storage and / or bandwidth requirements low. %
We place the sensor setup on a highway overpass ensuring an unobstructed view over a highway entrance segment from a vantage point. %
The unobtrusive placement avoids an influence on the observed driving behavior by not attracting any attention. %

%
\begin{figure}
    \centering
    \includegraphics[trim=75 146 75 90, width=1.0\columnwidth]{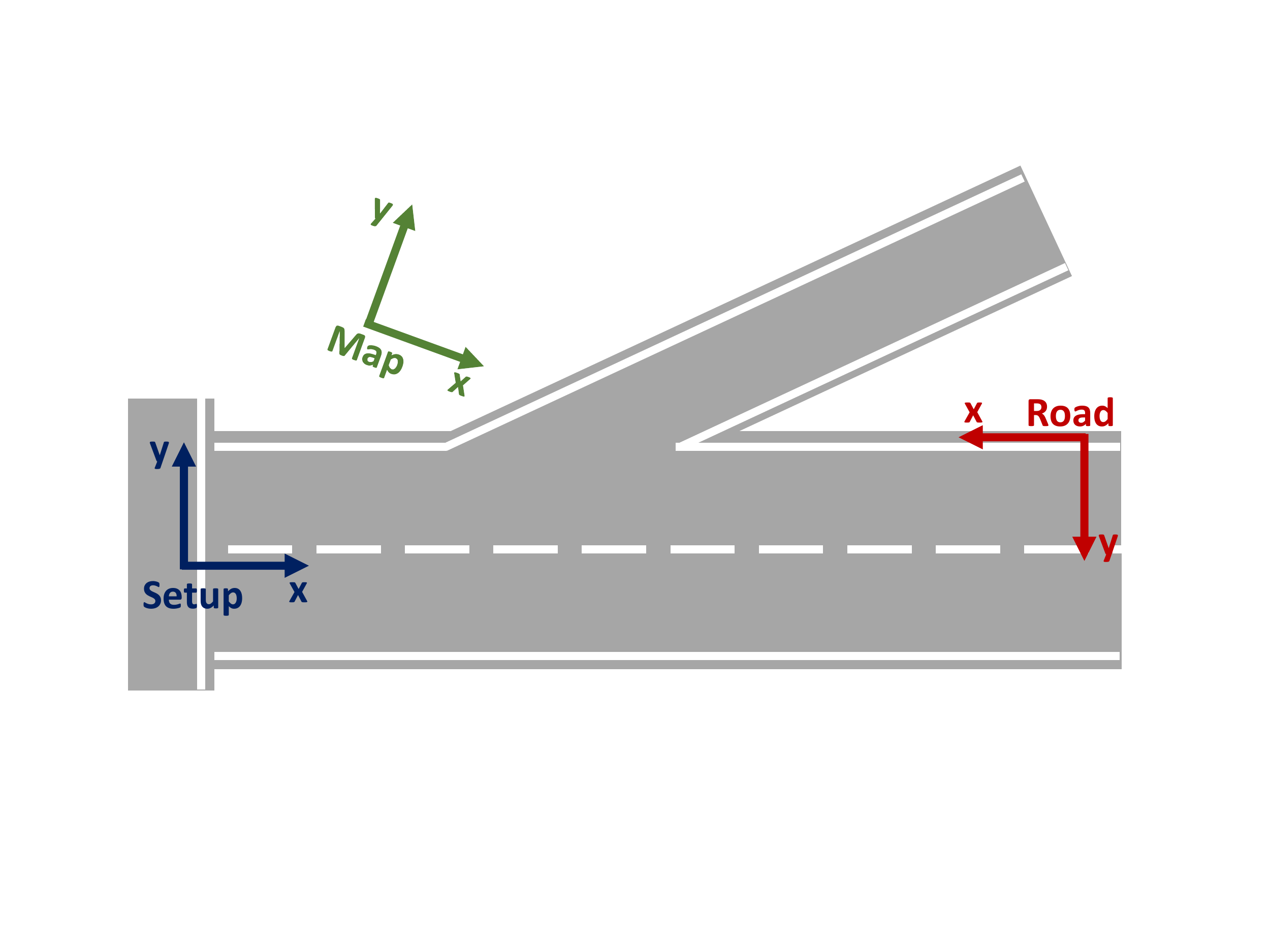}
    \caption{Visualization of the different coordinate systems used. The \textit{Setup} coordinate system is defined by the sensor placement and orientation. The \textit{Road} coordinate system is defined by the road points during the homography estimation. The projected map information is given in \textit{Map} coordinates.}
    \label{fig:coordinate_systems}
\end{figure}
Figure \ref{fig:coordinate_systems} visualizes the different coordinate systems used. %
Trajectories are recorded in the \textit{Setup} coordinate system defined by the sensor placement. %
However, they can be transformed into the \textit{Map} coordinate system as defined by a digital map file that contains a description of the static environment. %
For our experiments we use a map given in the OpenDrive format \cite{opendrive}. %
Object locations in the camera image are first transformed into the \textit{Road} coordinate system via a homography before being further transformed into the \textit{Setup} coordinate system. %
The homography estimation process will be described in more detail in the following section. %

\subsection{Camera Data Processing}
\label{subsec:camera}
Since we develop our own sensor setup we do have access to not only the camera images but also the camera itself. %
This access to the camera allows us to calibrate it with a checkerboard using Zhang's method \cite{zhang2000flexible} before the actual traffic recording. %
We then make use of the camera calibration output, the intrinsic camera matrix and distortion coefficients, to undistort and rectify the recorded images online. %
Subsequently, we use YOLOv3 \cite{redmon2018yolov3} to detect objects of the classes \textit{car}, \textit{bus}, \textit{truck} and \textit{motorcycle}. %
Among the best performing deep neural networks for object detection YOLOv3 yields a good trade-off between accuracy and inference time. %
We download pre-trained network weights for YOLOv3. %
To further reduce the inference time and be able to run the network inference in real-time on recorded images we use TensorRT \cite{tensorrt} for network optimizations and weight quantizations. %
With TensorRT we are able to achieve a speedup factor of about $3$ without a significant accuracy degradation. %
Finally, we filter out object detections outside of a predefined area in order to discard TOs driving in the opposite direction and potential false positives. %
The final object bounding boxes on a camera image are visualized in Figure \ref{fig:det_traj_image}.

\begin{figure}
    \begin{subfigure}{.48\columnwidth}
        \centering
        \includegraphics[width=1.0\columnwidth]{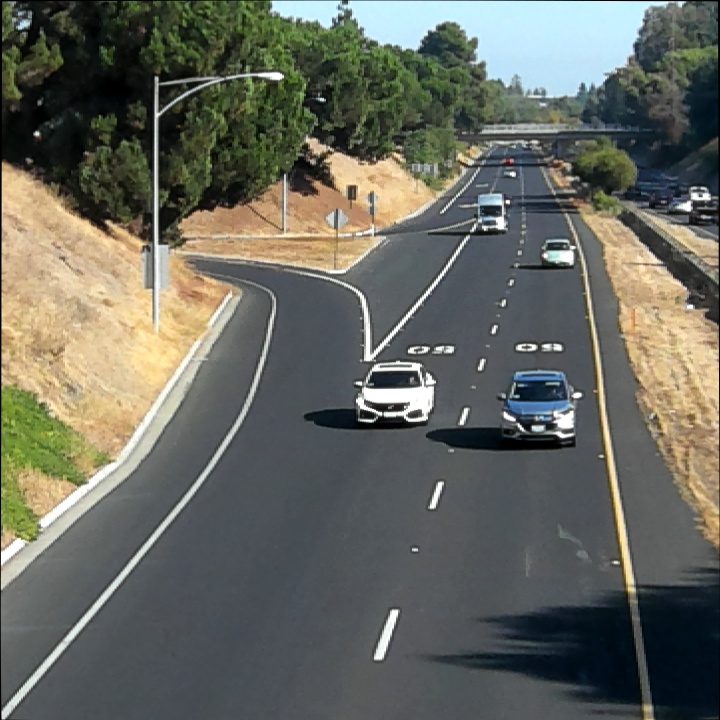}
        \caption{Image in camera plane.}
        \label{fig:homography_pre}
    \end{subfigure}
    \hspace{0.15cm}
    \begin{subfigure}{.48\columnwidth}
        \centering
        \includegraphics[width=1.0\columnwidth]{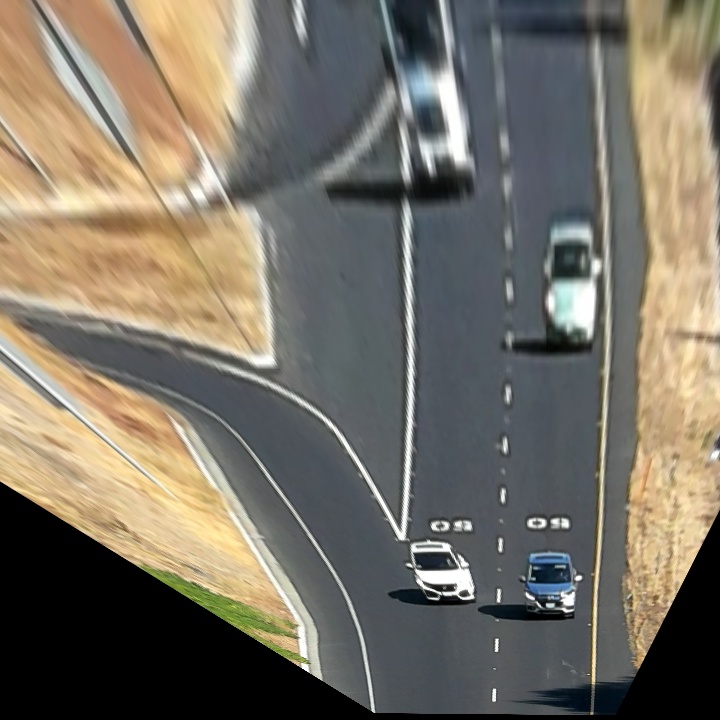}
        \caption{Image in road plane.}
        \label{fig:homography_post}
    \end{subfigure}
    \caption{Visualization of the estimated homography between the camera and the road plane with an image. %
    (a) shows an image as recorded by the camera. %
    (b) shows its transformation with the homography matrix into the road plane. %
    In this view, equal pixel distances in $x$ and $y$ correspond to equal distances on the road.}
    \label{fig:homography}
\end{figure}
Next, we need to transform the detected object locations from image coordinates into actual positions on the road. %
For that, we compute a homography \cite{chum2005geometric} between the image plane and the road, which we approximate by a plane. %
Homography estimation requires at least four point correspondences from the two planes. %
We select several features, such as beginnings and endings of lane markings, that can easily be detected on both the camera image and either a corresponding Google maps satellite image or our digital map. %
The points on the road are given in the \textit{Road} coordinate system as defined in Figure \ref{fig:coordinate_systems}, which is serving trajectory visualization purposes. %
Figure \ref{fig:homography} shows an exemplary image in image coordinates and its corresponding transformation with the homography matrix in road coordinates. %

We are able to further transform the \textit{Road} coordinates into the \textit{Setup} coordinate system. %
This transformation can easily be computed from the known or measured sensor setup location and orientation and the defined road coordinate system. %
A transformation into \textit{Map} coordinate system can again be computed by a set of corresponding points in \textit{Map} coordinates and any other coordinate system. %
Hence, object locations and trajectories can be transformed into different coordinate systems that serve special purposes. %

The computed homography can be used to transform an image point on the road surface into \textit{Road} coordinates. %
We compute and transform the lower front center point of the vehicles. %
This point is approximately given by center of the lower bounding box edge. %
An error is introduced when TOs drive at an angle towards the camera and the objects' sides are visible. %
We geometrically compensate for this error by using driving direction information from the object's history. %

\subsection{Fusion with Radar and Tracking}
\label{subsec:radarfusion}
Next to the camera, we use a state-of-the-art traffic surveillance radar. %
Once it is installed this radar executes a self-calibration based on background information and recorded traffic data. %
The radar performs target filtering, clustering and object detection internally and returns object lists in the \textit{Setup} coordinate system. %
Whereas each camera detection only contains the object's class, location and width, the radar also outputs its velocity, length and height. %

To benefit from the advantages of both camera and radar regarding object detection we perform a high-level sensor fusion and tracking on the detected object lists. %
Each object has its own track. %
The goal of our fusion and tracking algorithm is to assign all detections of an object to its respective track. %
Since the update rates of the camera and radar differ we do not synchronize them or fuse their detections into one coherent representation. %
Rather, we assign detections from both modalities to the same track and take their respective strengths into consideration during post-processing. %

We follow the tracking-by-detection paradigm and use the IoU tracker of Bochinski et al. \cite{bochinski2017high} as the basis for our algorithm. %
The IoU tracker operates in image coordinates and assigns a detected object to the track with whose last detection it has the highest IoU, as long as the IoU is above a minimum threshold. %
The IoU tracker is fast, works reasonably well and does not rely on stored image data. %
However, false negative detections, i.e. an actual object is not detected in a frame, lead to track fragmentation and identity switches. %
We are able to successfully solve this problem by keeping tracks alive for some time despite no new detections have been assigned. %
In our case, a keep-alive-threshold of \SI{0.5}{\second} works well. %
Furthermore, we complement the matching based on IoU in image coordinates with a matching based on Kalman filter predictions in \textit{Road} coordinates. %
Each track has an assigned Kalman filter with a constant velocity model that predicts the object's position and velocity at the time of the new measurement and matches new detections to tracks based on the nearest neighbor principle. %
In the discussed case of missed detections, the Kalman filter can extrapolate an object position and still correctly assign a new detection later. 
The state of an object is given by its state vector $\left( x^R, y^R, v_x^R, v_y^R, x_1^I, y_1^I, x_2^I, y_2^I \right)$. %
$\left(x^R, y^R \right)$ and $\left(v_x^R, v_y^R \right)$ correspond to position and velocity in \textit{Road} coordinates, and $\left(x_1^I, y_1^I\right)$ and $\left(x_2^I, y_2^I\right)$ define two opposite bounding box coordinates in the image. %
The Kalman filter only operates on the \textit{Road} values $\left( x^R, y^R, v_x^R, v_y^R \right)$. %
New camera detections update the road position estimates, whereas new radar detections also update the road velocity estimates. %
For camera measurements we choose a high uncertainty in $x$ and a low uncertainty in $y$, whereas for radar measurements we choose a low uncertainty in $x$ and a medium uncertainty in $y$. %
These measurement covariance matrices make effective use of each sensor's respective strengths. %
Due to the high measurement frequency our constant velocity motion model is quite accurate and we assume a low process noise. %

New radar measurements are matched to tracks based on $\left( x^R, y^R \right)$ via Kalman filter. %
New camera measurements are matched to tracks based on $\left( x^R, y^R, x_1^I, y_1^I, x_2^I, y_2^I \right)$ via both Kalman filter and IoU. %
%
%
If a currently detected object cannot be matched to an existing track due to an insufficient IoU overlap and / or a too large distance in the \textit{Road} space to the Kalman filter predictions a new track is created. %
Overall, this tracking algorithm performs very well in practice. %
In our visual examinations neither track fragmentation nor identity switches occurred. %
\subsection{Post-Processing}
\label{subsec:post-processing}
\begin{figure}
    \centering
    \includegraphics[trim=65 0 100 10, clip, width=1.0\columnwidth]{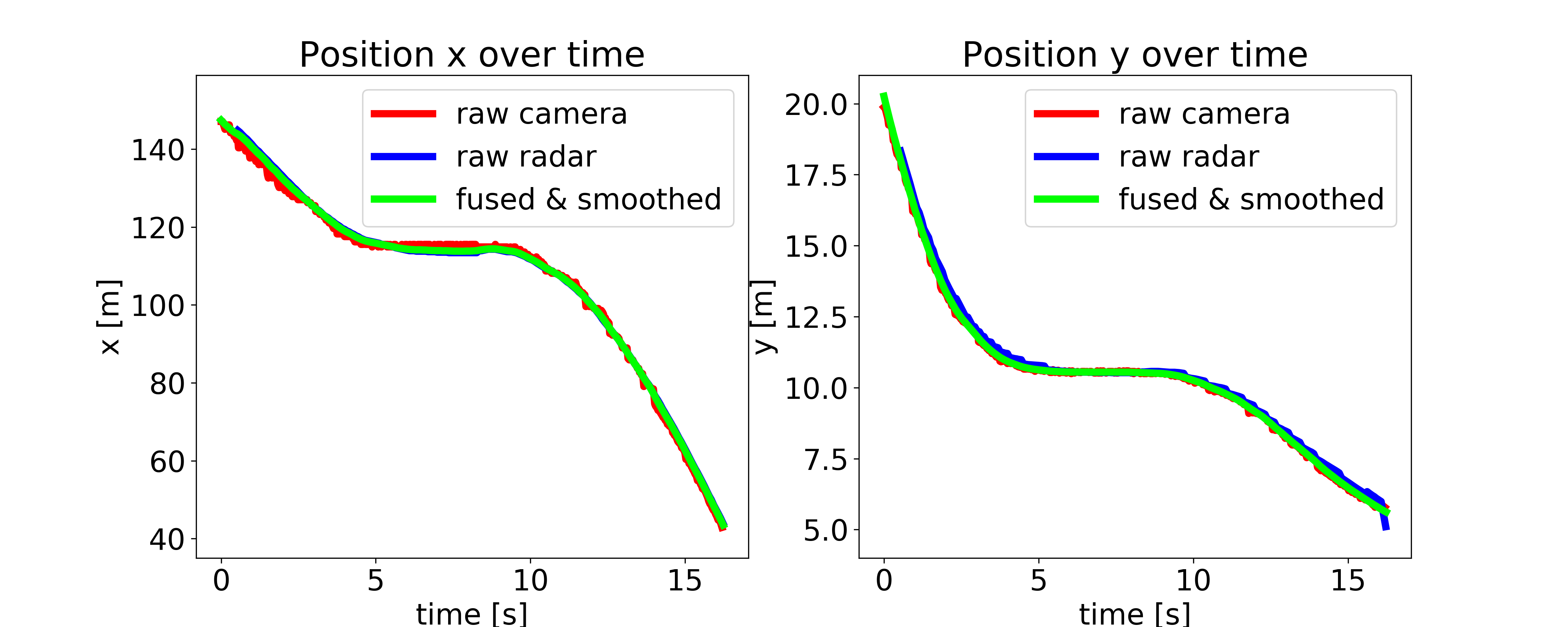}
    \caption{Raw camera and radar position measurements of a track as well as their fused and smoothed measurements over time. The final, RTS-smoothed estimates of the longitudinal position $x$ and the lateral position $y$ follow more strongly the radar and the camera measurements, respectively. %
    This behavior is due to our Kalman filter design. 
    }
    \label{fig:smoothing}
\end{figure}
Once tracks are completed we post-process them to smooth their noisy single measurements. %
For that, we apply the RTS smoother \cite{rauch1965maximum} to all detections of a track. %
The RTS smoother performs a forward Kalman filter pass followed by a backward Kalman filter pass. %
Hence, it effectively uses the fact that for finished trajectories both future and past information can be used for smoothing. %
Figure \ref{fig:smoothing} visualizes the smoothing effect. %
The resulting trajectory is less noisy than the raw measurements. %
Furthermore, the longitudinal position $x$ follows more strongly the radar measurements whereas the lateral position $y$ follows more strongly the camera measurements. %
The reason for this behavior is our choice of weights in the Kalman filter. %
We use different covariance matrices in the update-step for camera and radar measurements to put more weight on their respective strengths. %

\begin{figure}
    \centering
    \includegraphics[width=1.0\columnwidth]{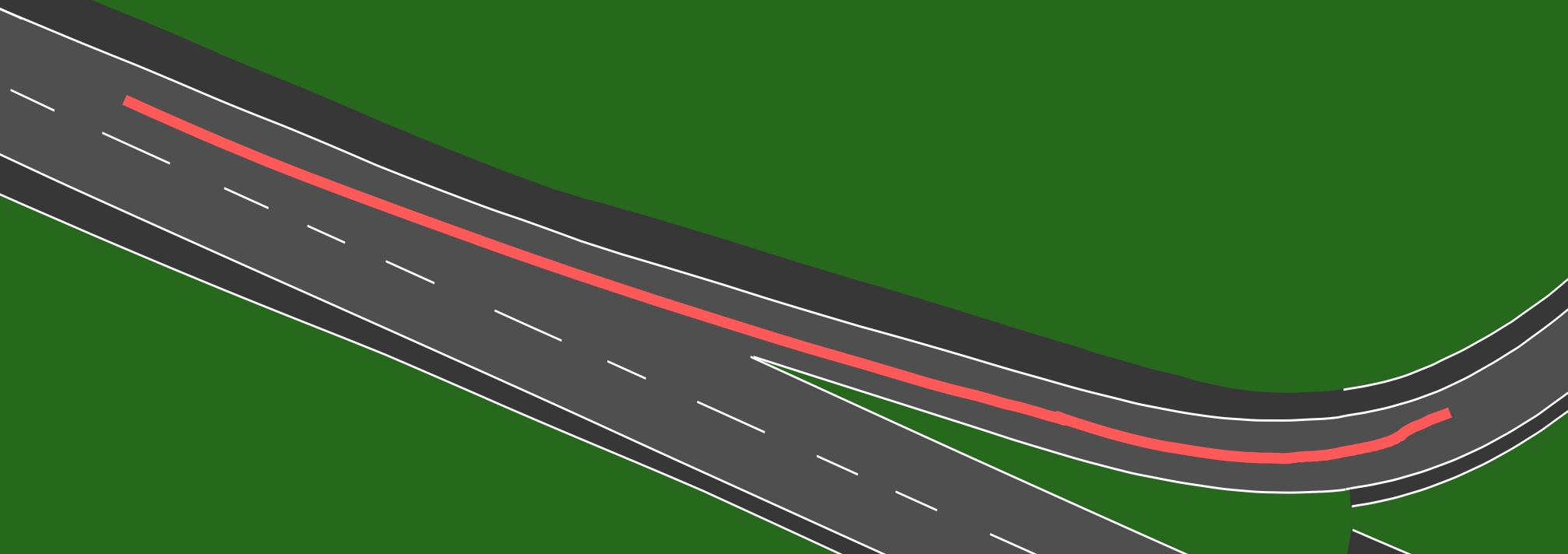}
    \caption{An extracted vehicle path overlayed on an OpenDrive map. The red path on the map corresponds to an extracted and smoothed trajectory of a vehicle entering the highway.}
    \label{fig:traj_on_map}
\end{figure}
Furthermore, for each track we determine a single object class and object dimensions by aggregating the information from all detections. %
We filter out the most extreme values and compute a weighted mean with more weight on measurements that occur closer to the sensor setup. %
Finally, each track consists of an object class, object dimensions and a list of detections, which contain timestamps, positions, velocities and heading angles. %
We are able to convert the final trajectories into \textit{Map} coordinates and visualize them on the map. %
An example is given in Figure \ref{fig:traj_on_map}.

\section{Experiment Design}
\label{sec:4_experiments}
\begin{figure}
    \centering
    \includegraphics[trim=0 12 0 100, clip, width=1.0\columnwidth]{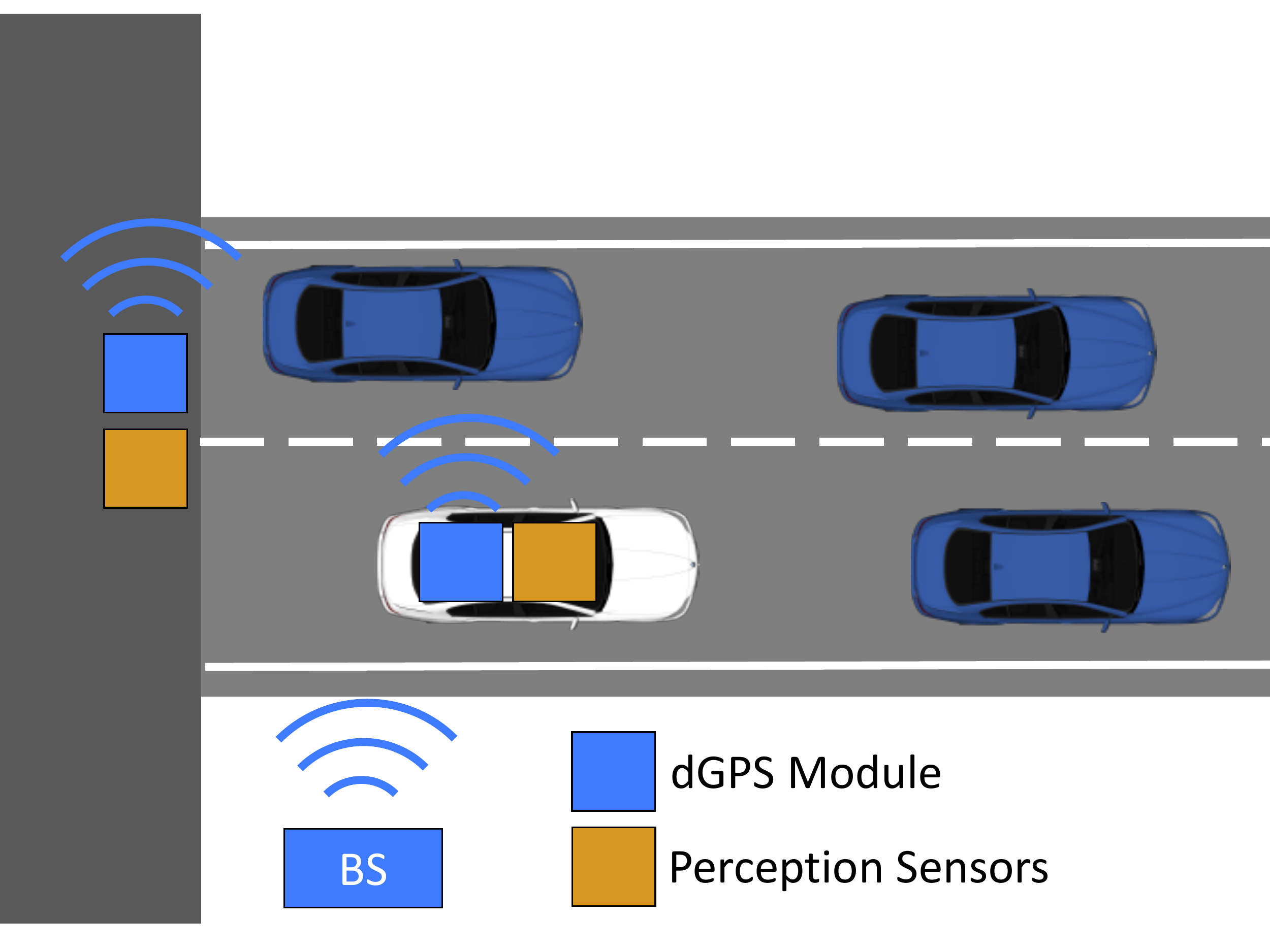}
    \caption{Our experiment setup for ground truth data generation. We equip both the infrastructure sensor setup and our own test vehicle with dGPS modules. We extract ground truth vehicle trajectories and convert them into the \textit{Setup} coordinate system. Then, we compare the ground truth data to the infrastructure detections and assess the accuracy of the extracted data.}
    \label{fig:overpass_setup}
\end{figure}

In the following, we describe our experiments for evaluating the accuracy of the trajectory extraction setup. %
In general, it is not feasible to acquire accurate ground truth TO trajectories through manual labeling. %
Particularly, for crowded scenes and TOs far away from the sensor setup human labelers cannot accurately label the position of objects. %
Hence, we decide to integrate a dGPS sensor into our vehicle and use it to record ground truth trajectories. %
Furthermore, we also integrate a dGPS sensor into the infrastructure sensor setup. %
This additional sensor allows us to easily compute the transformation between the moving vehicle and the fixed \textit{Setup} coordinate system. %
Figure \ref{fig:overpass_setup} shows a schematic overview of our experiment setup. %
%

We make use of the Network Time Protocol (NTP) to synchronize the clocks of our vehicle and infrastructure sensor setup. %
For an intercontinental Internet path, NTP achieves an accuracy of several tens of milliseconds \cite{mills2016computer}. %
In our case, the time synchronization accuracy should be better since we used an NTP server close the recording location. %
At a speed of \SI{50}{mph} \SI{10}{\milli\second} would still correspond to about \SI{22}{\centi\meter} in inaccuracy in the driving direction. %
A second source of error is the accuracy of the dGPS signal itself. %
With the obtained signals we achieve a localization accuracy of about \SI{20}{\centi\meter}.

We drive several times with our test vehicle through the area covered by our infrastructure sensor setup on both the highway entry lane and the two straight lanes. %
For each run, we extract only the trajectory of our vehicle from the infrastructure recordings to avoid having to solve a trajectory association problem. 
For the actual evaluation, we transform the reference (ground truth) trajectory into the \textit{Setup} coordinate system. %
We receive dGPS measurements with \SI{100}{Hz}, and we linearly interpolate between measurements from the infrastructure setup with the two closest timestamps. %
Then, we compute the error $p^{er}(d, i)$ at distance $d$ from the sensor setup for run $i$ as the difference of reference and measured values:
%
%
\begin{equation}
p^{er}(d, i) = p^{ref}(d, i) - \left. p^{me}(i)\right\vert_{t=t(p^{ref}(d, i))}
\label{eq:error}
\end{equation}
where $p \in \{x, y, v_{x}, v_{y}, \theta\}$ is a placeholder for the $x$ and $y$ position, the velocity in $x$ and $y$ direction and the heading angle. %
As shown in Equation \ref{eq:error}, we compare the reference values at a specific distance with the (interpolated) measurement values from the same time. %
We compute the errors for $d \in [\SI{35}{\meter}, \SI{135}{\meter}]$ since a vehicle entering the highway becomes visible at a distance of just over \SI{135}{\meter} and stops being fully visible at just under \SI{35}{\meter}. %
For each reference distance, we can then compute the mean error (bias) and the standard deviation over all of the runs:
\begin{equation}
\mu(p^{er}, d) = \frac{1}{N} \sum_{i=1}^{N} p^{er}(d, i) 
\label{eq:mu}
\end{equation}
\begin{equation}
\sigma(p^{er}, d) = \sqrt{\frac{1}{N} \sum_{i=1}^{N} \left [p^{er}(d, i) - \mu(p^{er}, d) \right]^{2}}
\label{eq:std}
\end{equation}
\section{Results}
\label{sec:5_results}
%
We have recorded a total of $10$ trajectories with our dGPS-equipped BMW 7 Series on a highway entry in California. %
The recordings were made on a sunny day such that we can evaluate our whole trajectory extraction pipeline, including the sensor fusion by comparing camera-only, radar-only and fused measurements. %
Furthermore, we recorded the general traffic for several hours in the afternoon. %
These trajectories contain actual realistic human driving behavior and we can use them to compute some meaningful statistics. %
On average, we recorded $20$ TO trajectories per minute with about $300$ object detections each. %
Objects going straight on the highway are detected first by the radar at distances of \SI{200}{\meter} and above, which is helpful for initializing the Kalman filter with a velocity estimate. %
The average recorded YOLOv3 detection confidence is $0.91$ and the average IoU for matched bounding boxes is $0.89$. %
Within one track an object has never been consecutively not detected for more than \SI{0.3}{\second}. %

%
\begin{figure}
    \begin{subfigure}{1.0\columnwidth}
        \centering
        \includegraphics[trim=10 40 10 0, width=1.0\columnwidth]{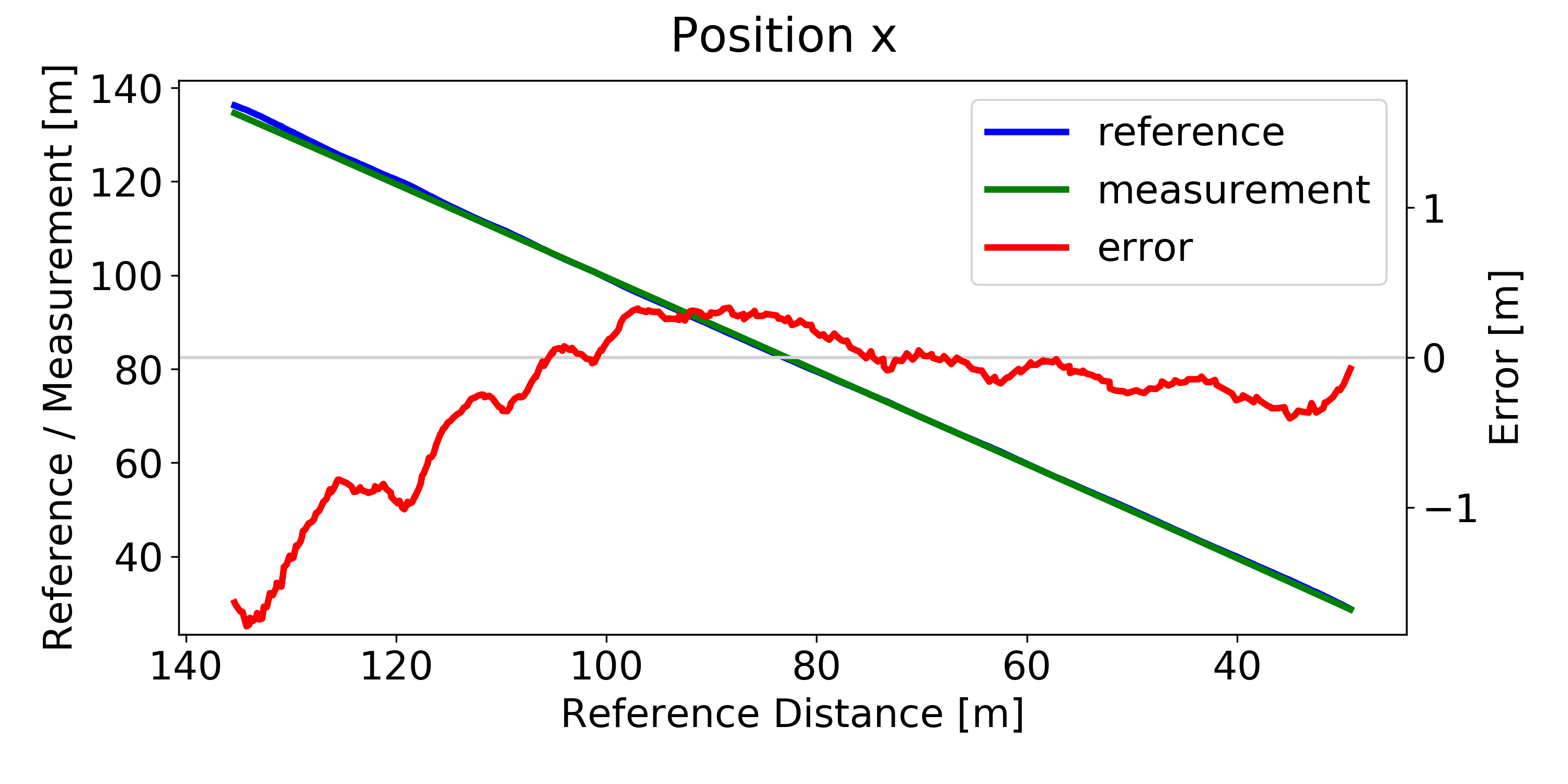}
        \label{fig:error_run3_x}
    \end{subfigure}
    \begin{subfigure}{1.0\columnwidth}
        \centering
        \includegraphics[trim=10 55 10 0, width=1.0\columnwidth]{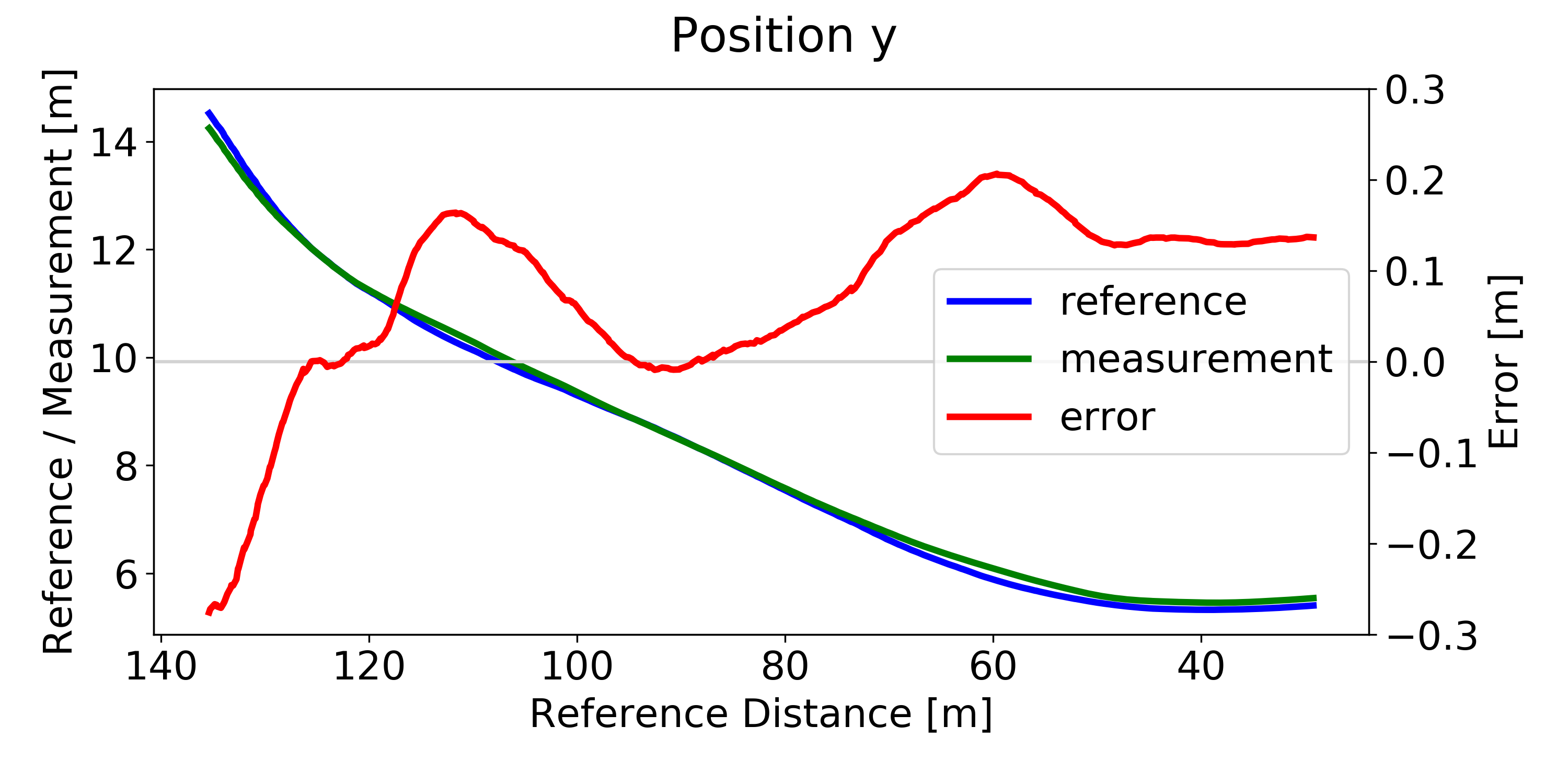}
        \label{fig:error_run3_y}
    \end{subfigure}
    \caption{Reference, measurement and error values for the position for a \textit{highway entry run}. %
    The graphs show the reference and infrastructure measurement values for position $x$ (top) and position $y$ (bottom) in the \textit{Setup} coordinate system on the left ordinate axes and the corresponding error values on the right ordinate axes. %
    For a reference distance of up to \SI{110}{\meter} the maximum absolute error in $x$ is less than \SI{0.5}{\meter}. %
    For a reference distance of up to \SI{130}{\meter} the maximum absolute error in $y$ is less than \SI{0.2}{\meter}. %
    }
    \label{fig:error_run3}
\end{figure}
\begin{figure}
    \begin{subfigure}{1.0\columnwidth}
        \centering
        \includegraphics[trim=10 40 10 0, width=1.0\columnwidth]{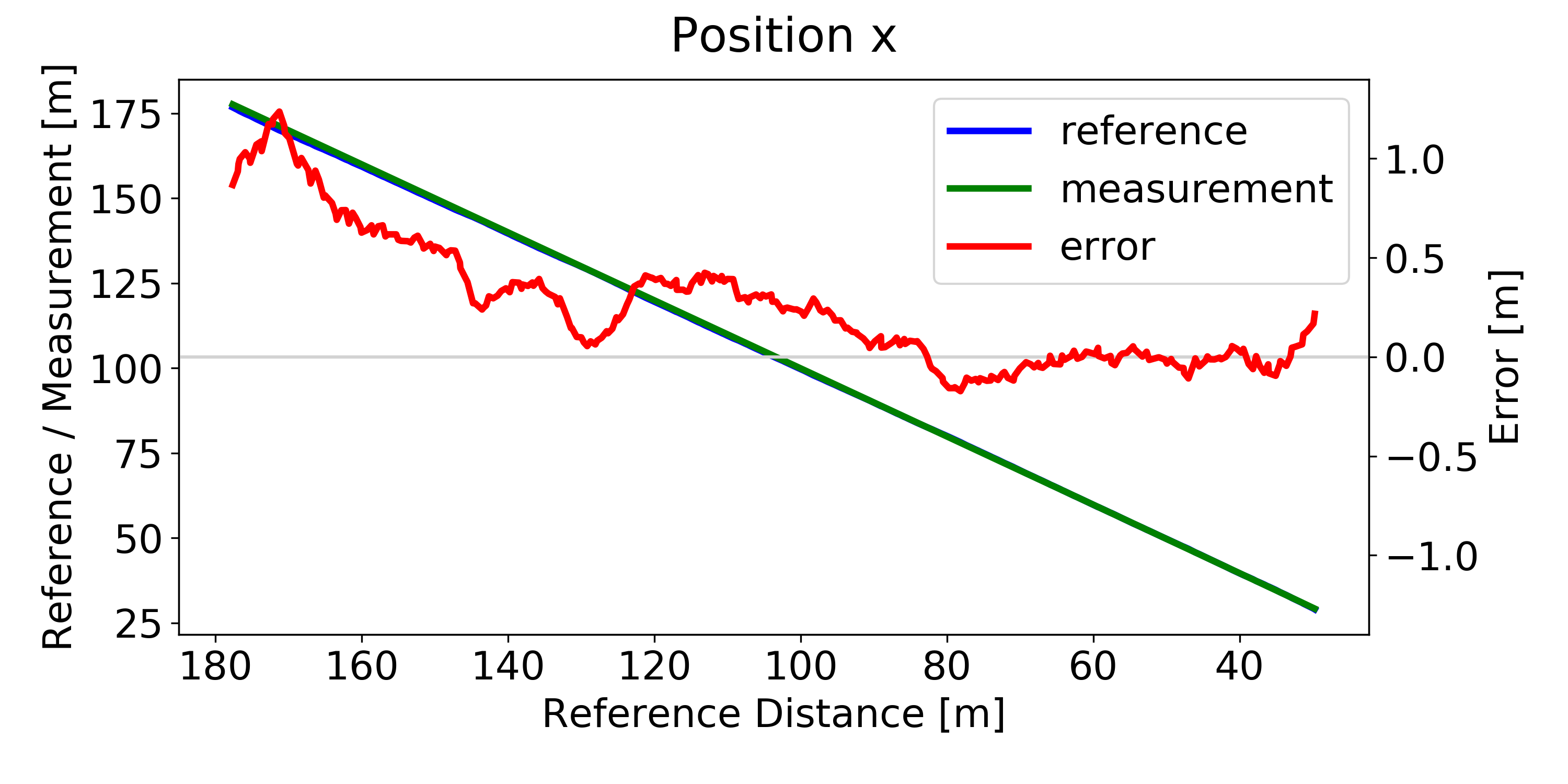}
        \label{fig:error_run6_x}
    \end{subfigure}
    \begin{subfigure}{1.0\columnwidth}
        \centering
        \includegraphics[trim=10 55 10 0, width=1.0\columnwidth]{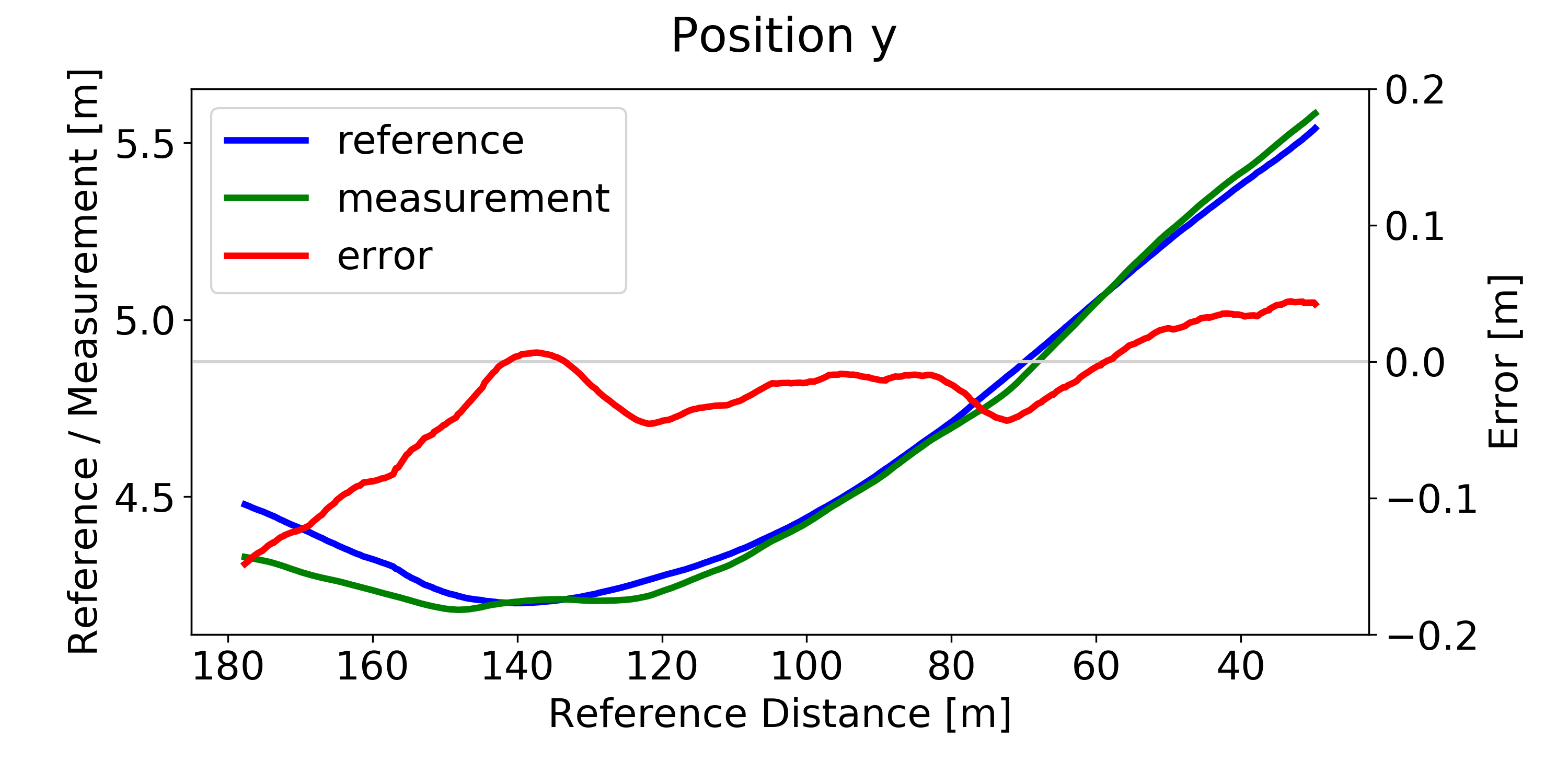}
        \label{fig:error_run6_y}
    \end{subfigure}
    \caption{Reference, measurement and error values for the position for a \textit{straight run} on the highway. %
    The graphs show the reference and infrastructure measurement values for position $x$ (top) and position $y$ (bottom) in the \textit{Setup} coordinate system on the left ordinate axes and the corresponding error values on the right ordinate axes. %
    For a reference distance of up to \SI{150}{\meter} the maximum absolute error in $x$ is less than \SI{0.5}{\meter}. %
    For a reference distance of up to \SI{160}{\meter} the maximum absolute error in $y$ is less than \SI{0.1}{\meter}. %
    Note that we plot the values for reference distances of up to \SI{180}{\meter} compared to up to \SI{135}{\meter} for highway entries. %
    }
    \label{fig:error_run6}
\end{figure}
Figure \ref{fig:error_run3} and Figure \ref{fig:error_run6} show plots for a single \textit{highway entry run} and a single \textit{highway straight run}, respectively. %
We plot the reference (dGPS) position values in blue and the position measurements from our infrastructure sensor setup in green against the reference distance of the object. %
These values are shown in the left ordinate axes. %
The corresponding error values, computed according to Equation \ref{eq:error}, are plotted in red and shown on the right ordinate axes. %
For the highway entry run we plot the values for reference distances in $[\SI{35}{\meter}, \SI{135}{\meter}]$, whereas for the highway straight run we plot the values for reference distances in $[\SI{35}{\meter}, \SI{180}{\meter}]$. %
Overall, for these single runs, the errors are, even for large distances, very low. %
For instance, for the highway straight run the maximum absolute error in $x$ stays below \SI{0.5}{\meter} for distances up to \SI{150}{\meter} and the maximum absolute error in $y$ stays below \SI{0.1}{\meter} for distance of up to \SI{160}{\meter}. %

%
\begin{figure}
    \begin{subfigure}{1.0\columnwidth}
        \centering
        \includegraphics[trim=10 40 10 0, width=1.0\columnwidth]{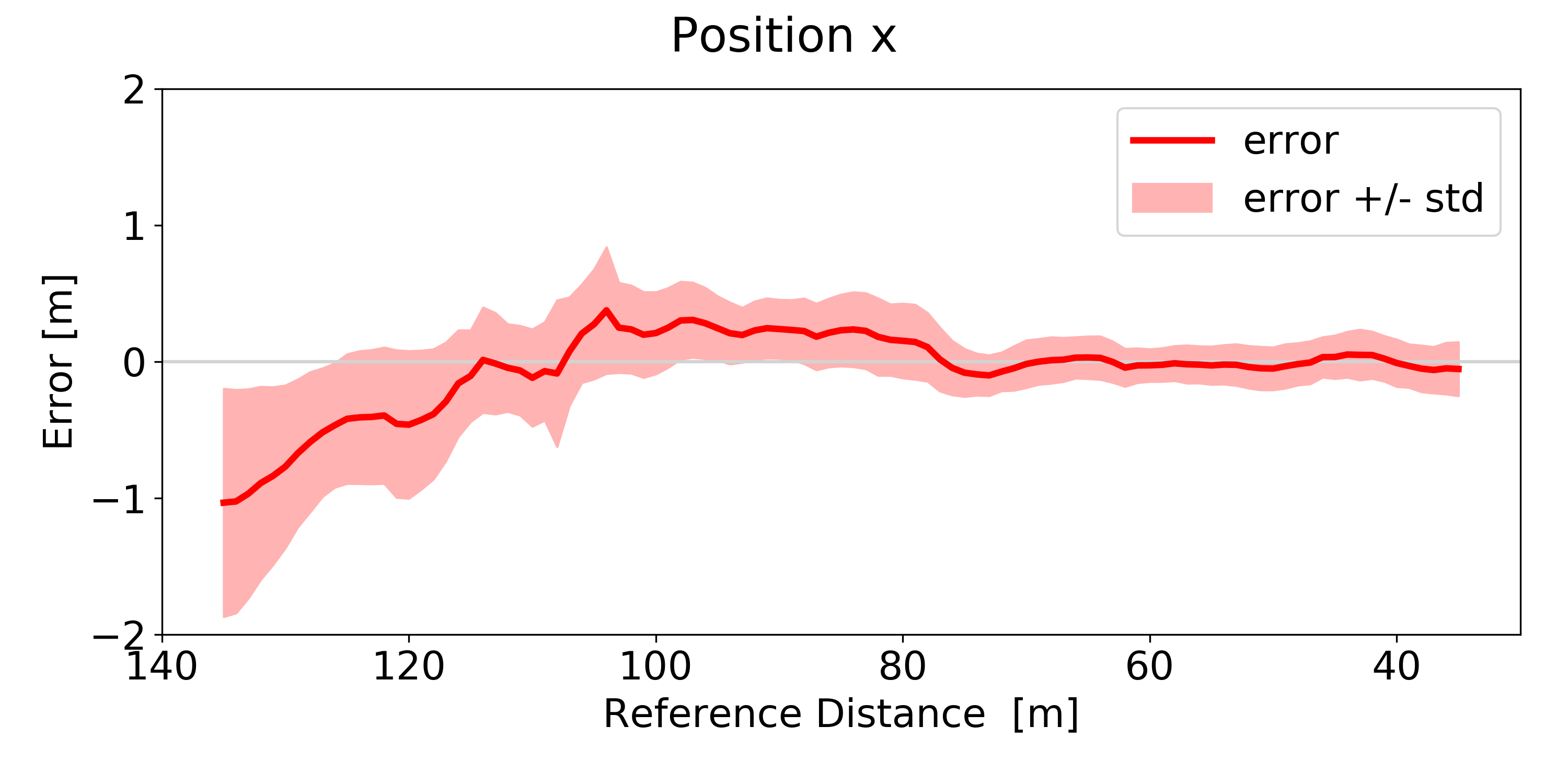}
        \label{fig:error_agg_x}
    \end{subfigure}
    \begin{subfigure}{1.0\columnwidth}
        \centering
        \includegraphics[trim=10 55 10 0, width=1.0\columnwidth]{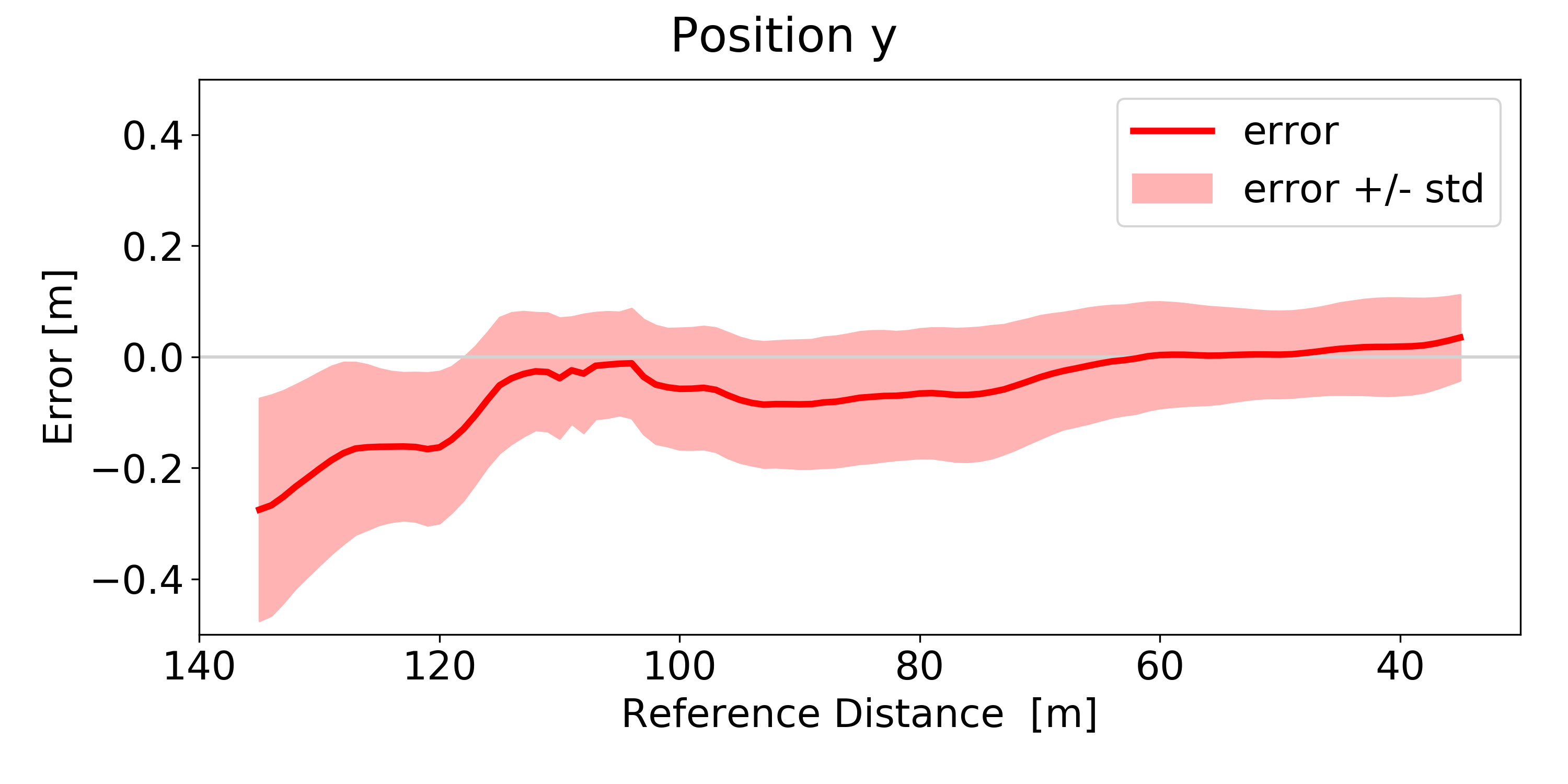}
        \label{fig:error_agg_y}
    \end{subfigure}
    \caption{Average positional measurement error and its standard deviation against the reference distance of the infrastructure sensor setup to the vehicle. %
    The mean is taken over $10$ runs. %
    For position $x$ (top) and position $y$ (bottom) in the \textit{Setup} coordinate system the mean error is plotted in red and surrounded by one standard deviation in each direction plotted in light, transparent red. %
    The error uncertainty increases with the reference distance. %
    However, both error and uncertainty are very low for up to \SI{120}{\meter}. %
    }
    \label{fig:error_agg}
\end{figure}
Next to single trajectory analyses, we also look at the aggregated errors from the $10$ runs. %
To challenge our trajectory extraction pipeline, we recorded the majority of the runs, $7$, from highway entries and the rest, $3$, from straight drives on the highway. %
In these runs, we accelerated and braked on purpose to avoid a simple constant velocity driving. %
We compute both mean and standard deviation of the error against the reference distance according to Equations \ref{eq:mu} and \ref{eq:std} for the interval $[\SI{35}{\meter}, \SI{135}{\meter}]$. %
Figure \ref{fig:error_agg} visualizes the mean error and its corresponding standard deviation from the $10$ runs for the position $x$ and $y$. %
We can see from the plots that the standard deviations of the errors increase with an increasing reference distance, which confirms our expectations. %
Both error and standard deviation are very low, which proves the potential of infrastructure sensors to extract highly accurate trajectory data. %

%
\begin{table}
\centering
\small
\resizebox{\columnwidth}{!}{%
\begin{tabular}{|c|ccc|}
    \hline
    & \multicolumn{3}{c|}{\textbf{Mean Bias (Mean Standard Deviation)}}\\
    \cline{2-4}
    & Camera & Radar & Fused\\
	\hline
	$x$ $\left[ \si{\meter} \right]$ & $-0.56$ ($0.50)$ & $0.08$ ($0.29$) & $-0.06$ ($0.29$)\\
	$y$ $\left[ \si{\meter} \right]$ & $0.04$ ($0.10)$ & $-0.11$ ($0.24$) & $-0.06$ ($0.11$)\\
	$v_x$ $\left[ \si{\meter\per\second} \right]$ & $0.04$ ($0.51)$ & $0.06$ ($0.11$) & $0.08$ ($0.13$)\\
	$v_y$ $\left[ \si{\meter\per\second} \right]$ & $0.07$ ($0.10)$ & $0.09$ ($0.14$) & $0.09$ ($0.14$)\\
	$\theta$ $\left[ \si{\degree} \right]$ & $0.10$ ($0.49)$ & $0.33$ ($0.70$) & $0.33$ ($0.66$)\\
	\hline
\end{tabular}
}
\vspace{1mm}
\caption{Mean bias and mean standard deviation over the reference distance interval $[\SI{35}{\meter}, \SI{135}{\meter}]$ for camera-only, radar-only and fused tracks. %
The bias is of less importance than the standard deviation since this systematic error can be subtracted from the resulting trajectories. %
The fused tracks are as good as the camera in $y$  and as good as the radar in $x$, $v_x$ and $v_y$. %
Hence, the fusion makes good use of the respective strengths of both sensors.}
\label{tab:results}
\end{table}
Lastly, we want to compare the accuracy of the fused camera and radar data to camera-only and radar-only measurements from our vision pipeline. %
For that, we compute the mean of both mean error and standard deviation over reference distances in $[\SI{35}{\meter}, \SI{135}{\meter}]$. %
The results are displayed in Table \ref{tab:results}. %
Next to the errors in $x$ and $y$, we also compute errors for the velocities $v_x$ and $v_y$ and the extracted heading angles $\theta$. %
The mean error or bias is less of an interest since we can use the computed values to de-bias our measurements. %
However, we cannot reduce the mean standard deviation. %
From the results, we can see that our sensor fusion algorithm successfully combines the advantages of both sensors. %
In the lateral position $y$, the fused results follow the low camera errors, whereas in the longitudinal position $x$ and the velocities they follow the low radar errors. %
The heading angle $\theta$ is directly computed from the velocities and hence, also mainly influenced by the radar. %

As mentioned before, we can use the mean bias values against reference distance values to de-bias the data. %
This de-biasing corresponds to increasing the accuracy of the extracted trajectories with our test data. %
Since the standard deviations of the errors are very low our extracted trajectories are, even for large distances, highly accurate. %
The reason for this high accuracy, next to our advanced vision pipeline, is the advantage of infrastructure sensors over autonomous test vehicles: %
Unobstructed field of view from a vantage point and no self-localization errors. %
Furthermore, when analyzing the presented errors, the inaccuracy of the acquired ground truth data has to be taken into consideration. %
From the NTP time synchronization, we expect \SI{22}{\centi\meter} position inaccuracy (mainly in $x$). %
From the dGPS, the radial inaccuracy ($\sqrt{x^2+y^2}$) should be below \SI{20}{\centi\meter}. 
Hence, our positional error standard deviations are in the same magnitude as these ground truth data inaccuracies. %
\addtolength{\textheight}{0cm}

\section{Conclusion}
\label{sec:6_conclusion}
%
In this work, we present a novel approach for the extraction of realistic driving trajectories from infrastructure sensor data. %
The use of infrastructure sensors allows for the recording of the whole, unaffected traffic space. %
Furthermore, recording a lot of data of a specific scenario is easy and fast. %
Our trajectory extraction pipeline is based on sensor data from a camera and a traffic surveillance radar. %
We perform object detection, transformation, tracking, fusion and trajectory post-processing. %
Our tracking algorithm improves the state-of-the-art IoU tracker \cite{bochinski2017high} by keeping unmatched tracks alive and adding a Kalman filter in the \textit{Road} coordinate system. %
These improvements solve the problem of track fragmentation and id switches. %
While existing TO trajectory datasets lack accuracy estimates, we collect accurate ground truth data with our dGPS-equipped test vehicle and evaluate the accuracy of the extracted trajectories. %
We use the mean errors to de-bias the extracted trajectories. %
The remaining error standard deviation is with about \SI{29}{\centi\meter} in $x$ and \SI{11}{\centi\meter} in $y$  similar to the ground truth data inaccuracy. %
Hence, our extracted trajectories are highly accurate and demonstrate that infrastructure sensors are highly suitable for extracting TO trajectory data. 
%

The collected data can be used for a variety of purposes. %
As suggested in \cite{notz2019methods} models that try to imitate human driving behavior can be trained on our realistic TO trajectory data. %
Such models can be used in simulators and significantly increase their realism. %
More realistic simulators improve simulation-based AD function development and can be used for verification and validation of autonomous vehicles. %
Another direction of future research is using traffic data recorded by infrastructure sensors in vehicle-to-infrastructure communication scenarios to direct traffic online and increase road users' safety. %
Furthermore, we expect interesting insights from further analyses of the recorded behavioral data. %
Highly accurate infrastructure recordings from different locations, times and weather conditions will allow for analyzing the influence of these factors on the human driving behavior. %
Eventually, such insights will support the development and implementation of autonomous vehicles. %

\section*{ACKNOWLEDGMENT}
We would like to thank %
Anthony Acker, %
Andrew Dickens, %
Mehmet Inönü, %
Sebastian Kienitz, %
Florian Münch, %
and Brad Siedner %
for their support with the construction of the hardware prototype and the execution of the data recordings. %

{\small
\bibliographystyle{IEEEtran}
\bibliography{references}
}

\end{document}